\documentclass{article}
\usepackage{spconf,amsmath,graphicx}

\title{Modelling the Human Intuition to Complete the Missing Information in Images for Convolutional Neural Networks}
\twoauthors
 {Robin Koç \sthanks{The first author performed the work
	while at ...}}
	{
 Middle East Technical University\\
Computer Engineering\\
Ankara, Turkey\\
 }
 {Fatoş T. Yarman Vural \sthanks{The second author performed the work
	while at ...}}
	{
 Middle East Technical University\\
Computer Engineering\\
Ankara, Turkey\\
 }

\begin{document}
%
\maketitle
\begin{abstract}
In this study, we attempt to model intuition and incorporate this formalism to improve the performance of the Convolutional Neural Networks.

Despite decades of research, ambiguities persist on principles of intuition. Experimental psychology reveals many types of intuition, which depend on state of the human mind.

We focus on visual intuition, useful for completing missing information during visual cognitive tasks. First, we set up a scenario to gradually decrease the amount of visual information in the images of a dataset to examine its impact on CNN accuracy. Then, we represent a model for  visual intuition using Gestalt theory. The theory claims that humans derive a set of templates according to their subconscious experiences. When the brain  decides that there is missing information in a scene, such as occlusion, it instantaneously completes the  information by replacing the missing parts with the most similar ones.

Based upon Gestalt theory, we model the visual intuition, in two layers. Details of these layers are provided throughout the paper.

We use the MNIST data set to test the suggested intuition model for completing the missing information. Experiments show that the augmented CNN architecture provides higher performances compared to the classic models when using incomplete images.
\end{abstract}
\section{Introduction}
\label{sec:intro}

The contemporary Machine Learning methods, specifically deep models, are known to be highly nonlinear function approximators, which map very low-level representations of physical phenomena, to  high-level concepts \cite{C1}. Most of the  algorithms are based on the assumption that the probability density functions of the training and test sets are generated by the same source, hence, they have the same distribution. Furthermore, it is assumed that the data set constitutes statistically sufficient and balanced categories with disentangled attributes. The methods that do not meet these assumptions are inexplicably, unpredictably, and uncontrollably fail to fit a function, which maps the input images to categories.

In the Deep Learning literature, there are many studies trying to make the models explainable and interpretable. For example, the article titled "Visualizing and Understanding Convolutional Networks" by Simonyan and Zisserman \cite{C2} has performed an analysis of CNN models through visualization of intermediate steps. The study of Qin, Yu, Liu, and Chen \cite{C3} discusses visualization methods used to explain how CNN models "see" the world. Additionally, new samples similar to the ones in the training set can be generated with adversarial generative networks, known as GANs. Thus, it is attempted to contain possible examples that can be encountered during the test in the training set \cite{C4}. However, examples produced with GAN models introduce a great bias and lead to significant stability problems. A group of models have managed to highly deceive the model by changing only a few pixels in the images in the training set \cite{C5}, which can be taken as a sign of the unpredictable nature of deep architectures. 

A line of deep models attempts to complete the missing parts of images by incorporating the context. A survey on image completion is found in \cite{C9}, where traditional image completion methods and their limitations are discussed. A generative image completion model is introduced by Yu et.al. \cite{C10}, which introduces a cascaded network for impainting the images. In \cite{C11},  Liu et. al investigate  the use of deep generative models for semantic image inpainting by combining  global and local context for more coherent and contextually-aware completions.

Although there are many sophisticated solutions that have different standpoints on deep learning, none of the current models include a model for intuition. In   \cite{dideye},  we made a preliminary endeavor to formalize visual intuition. In this paper, we further extend and analyze previous findings by conducting thorough experiments and offering deeper insights for incorporating visual intuition in CNN architectures. 

Intuition is a crucial ability of the human brain to produce information. Although there is a strong effort to uncover the mysteries of our brain, we still don't know how intuition works. However, experimental psychology reveals that there are many types of intuition, which can be observed in a wide range of cognitive tasks, such as problem-solving, object recognition, prediction, and decision-making. In this study, we shall only focus on our ability to complete the missing "visual" information, which is proposed as a type of visual intuition in Gestalt theory. We suggest a mathematical representation for this narrowed definition of visual intuition for information completion ability.

\section{Generation of Incomplete Image Test Dataset}

In the first step of our study, we prepare a scenario to measure the sensitivity of a CNN  model to the missing parts  of  the images in the test sets. In this scenario, we generate multiple test sets with gradually increased information loss. 

Formally speaking, suppose that we have a test set containing a total of $N^{te}$ images, selected from $C$ categories:

\begin{equation}
(\mathbf{x}_i, y_j) \in D^{te},
\end{equation}
where each image $\mathbf{x_i}$ is associated with a label $y_j$, $j=,..,C$.
We generate  multiple test sets from $D^{te}$ by gradually decreasing the information content of the images,  in two steps, namely, segmentation and information reduction steps, as explained below.

\subsection{Segmentation}

We propose a simple segmentation algorithm, based on region-growing to partition each image, $(\mathbf{x}_i, y_j) \in D^{te}$, into a set of regions, as follows.

For every pixel $p$ in our image, let's denote the set of neighboring pixels at a distance of 1 unit based on the Chebyshev distance by $\eta(p)$. The Chebyshev distance in a two-dimensional space, for points $p_1$ and $p_2$ with Cartesian coordinates $(x_1, y_1)$ and $(x_2, y_2)$, is calculated as:

\begin{equation}
U_{Chebyshev} = max(|x_1 - x_2|,|y_1 - y_2|).
\end{equation}

If the Chebyshev distance  between any pixel in the image, $p \in \mathbf{x}_i\ , \ p \neq 0$, and its neighboring pixels, $p^{'} \in \eta(p)$, is less than a predefined threshold,  we add the neighboring pixel to the segment, where the selected pixel is located.

Finally, we modify the segmented image by adding segments smaller than a certain size to neighboring segments.

\subsection{Information Reduction}

In order to generate test sets with controlled information deficit from our original test set, consisting of segmented images, we randomly delete $s$ segments from each image to produce, what we call, deficient test data sets,

\begin{equation}
{(\mathbf{x}_i, y_j})_{S-s}\in D_s^{te}.
\end{equation}

\begin{figure}[t]
\centerline{\includegraphics[scale=0.09]{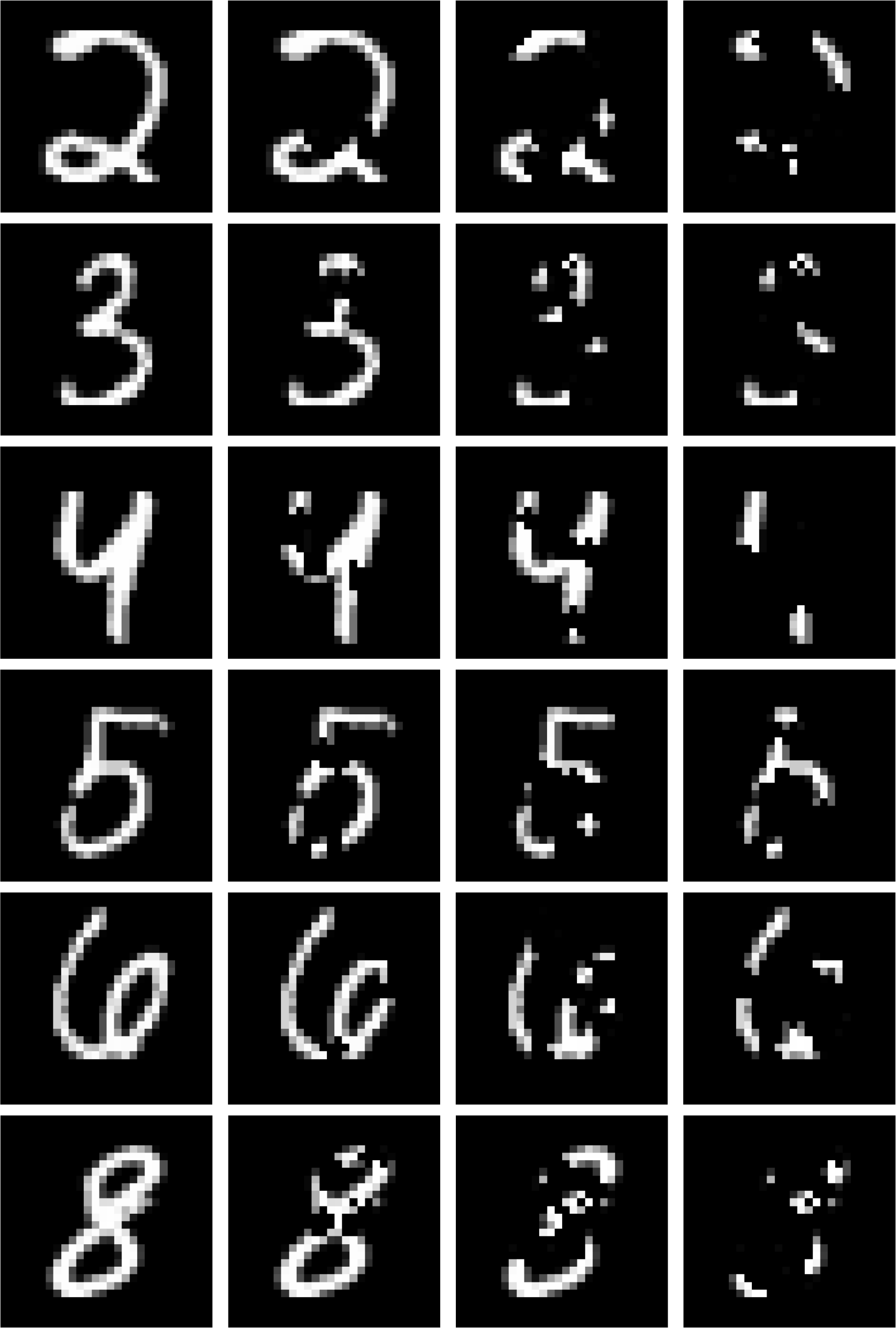}}
\caption{From left to right columns, sample images of digits 2, 3, 4, 5, 6, and 8 with $s=0, 4, 8$ and 16 deleted segments. The first column represents the complete images with $s=0$, used for training the CNN. The images in the rest of the columns, generated with  $s\ge 4$,  are the samples of   incomplete test sets with an increased information deficiency.}
\label{fig_info_lost}
\end{figure}

In the above formulation, $S$ indicates the total number of segments in an image and the subscript, $(S-s)$ indicates that out of $S$ segments, $s$ of them are deleted, where  $ s=1,...,S$ is the index representing the number of deleted segments from all of the images, $\mathbf{x}_i\in D^{te}$, to generate a  deficient test data set, $D_s^{te}$ . Figure \ref{fig_info_lost} shows samples from segment-reduced images of number digits, selected from the MNIST-Number data set. Notice that, as we keep removing  the number of segments, $s$, in the image,  we gradually lose the contextual information. Furthermore, for the values of $s\ge 4$, numbers start to resemble another digit than the original digit. For example, the number $8$ looks like the number $3$  and the number $5$ looks like the number $6$. Similar behavior is observed in the other number-digit images.  We generate  test sets by gradually decreasing the number of segments for $s= 1,....s_{max},$ where $s_{max}$ is an empirically set parameter, where the images are no further recognizable by human perception.

Thus, we obtain a total of $s_{max}$ deficient test data sets, consisting of images with gradually decreasing number of segments. These deficient test sets are included in the test dataset of Figure 2. We measured the performance of a simple CNN model, trained with the images of complete  MNIST images. Then, we measure the performance of this model by using the deficient test sets for $s= ,...,s_{max}$.  As we expected the performance of the CNN model gradually decreases as we delete more and more segments in the images. The performance decrease as a function of the deleted segments is shown by the blue curve in Figure \ref{fig_comparison}. As it can be observed from this figure, after deleting $s_{max} = 16$ segments from each image, the performances of the deficient test dataset get below the chance level.

\section{Algorithmic Thinking and  Intuitive Thinking Pathways}
Throughout the evolution of humanity,  our brain has adapted to meet essential cognitive requirements by engaging  information  through different cognitive pathways, which can be considered  as the \textbf{algorithmic thinking pathway} and the \textbf{intuitive thinking pathway}. These parallel cognitive processes contribute to the intricate and adaptive nature of human decision-making and problem-solving capabilities. The algorithmic thinking pathway makes inferences by processing the holistic information it perceives. The intuitive thinking pathway quickly completes the perceived missing information and transmits it to the algorithmic thinking pathway. 

 The inferences of the algorithmic thinking pathway are relatively slow but reliable.  They are tightly linked to information integrity and consistency. On the other hand, the intuitive thinking pathway is fast and comes into play in scenarios, where information is insufficient or misleading.

In this study, we assume that algorithmic thinking  corresponds to classical machine learning models, such as CNN. We attempt  to make this model robust with a simple and fast subconscious memory and intuition  layers, based on Gestalt theory. Although there are many theories about the human visual system, Gestalt theory provides us with an important clue to define visual intuition. According to this theory, \textbf{the missing information in the thought process of the human mind is quickly completed without the need for a certain inference process, based on our past experiences, stored  in our subconscious memory (Köhler, 1938)} \cite{C6}.

\begin{figure}[t]
\centerline{\includegraphics[scale=0.16]{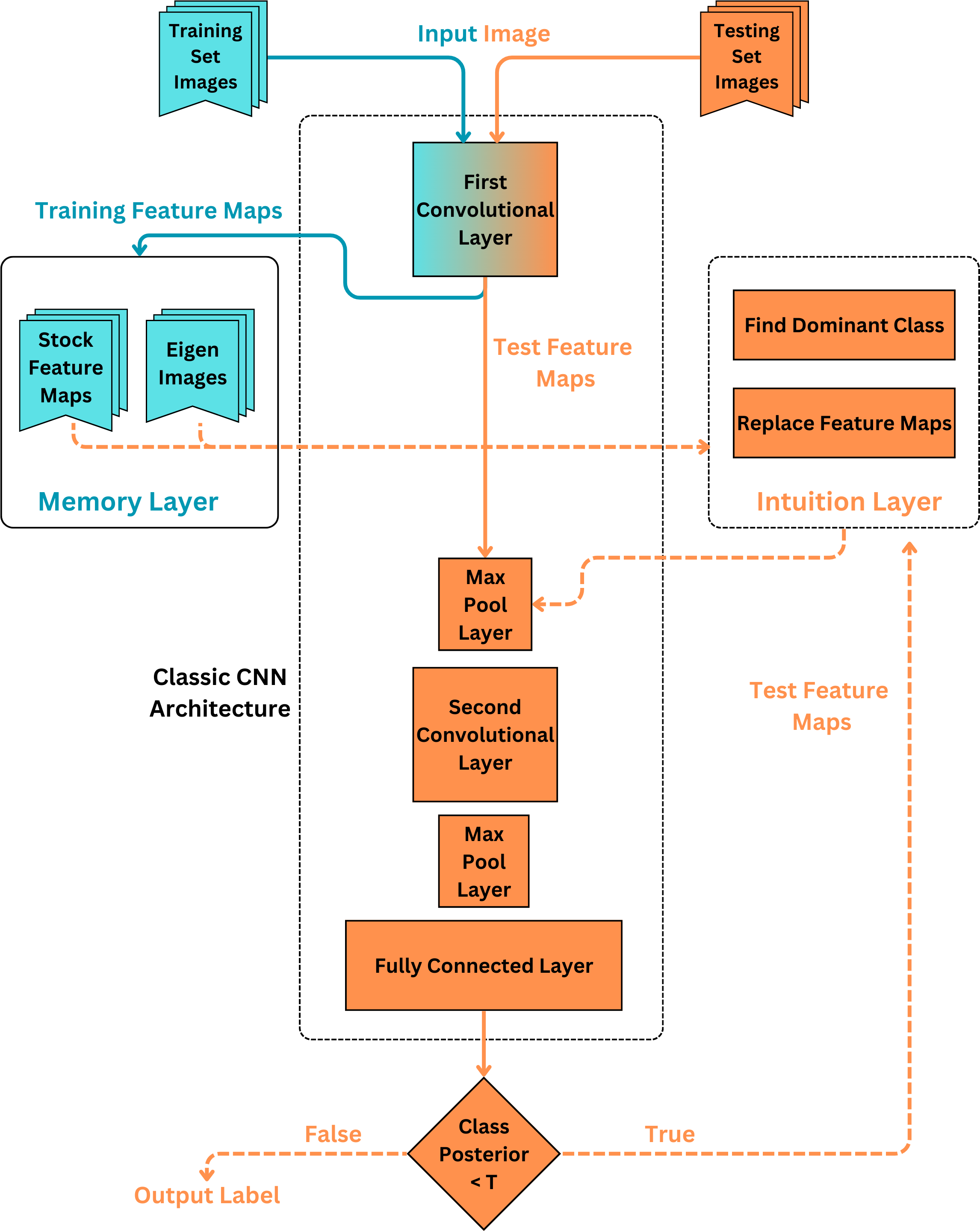}}
\caption{A simple CNN Model, augmented with subconscious memory and intuition layers: Orange color represents the test steps, augmented by the intuition layer and  blue color represents the training steps, augmented with  intuition layer. Dashed lines are activated only when the posterior probabilities of the fully connected layer, obtained in the test step, are below a predefined threshold value.}
\label{fig_schema}
\end{figure}
\section{An Augmented  CNN Architecture Using Subconscious Memory And Intuition layers}
Based on the Gestalt theory, we introduce a CNN architecture, augmented by  subconscious memory layer and  visual intuition layer (see: Figure 2). 

The training step of the standard CNN model is augmented by a memory layer, which represents the subconscious past experience.  During the training phase, we use  images with no information deficit to extract templates for each category. These templates, called the eigen images and stock feature maps, represent our subconscious memory and are stored in the memory layer. 

The test step of CNN is augmented by the intuition layer. We start the test step by  running the standard CNN model.  If the  posterior probability of  the estimated  category is below a threshold value,  the memory  and intuition layers are activated  and the dashed lines connect the standard CNN architecture to the memory and intuition layers. The activation of these layers  rapidly completes the missing information in the test image by retrieving the eigen images  to estimate the dominant class of features of the test image. Once the dominant class is estimated,  the feature map of the test image is replaced with the stocked feature map of the dominant class. Finally, the standard CNN  tests  the replaced feature maps to make the ultimate decision.  

During our experiments, we observed that the augmented CNN model is relatively more robust to incomplete information, compared to the classical CNN architecture. In the following subsections,  we make the formal definition of the suggested visual intuition model, which consists of a memory layer and an intuition layer. Then, we explain a CNN architecture, augmented by the suggested intuition model.  

\subsection {Memory layer: A Model for the Subconscious Past Experience with Template Images}

According to Gestalt theory,  objects exist in the human mind with  template models, based on individual experiences. These templates contain common features of different objects belonging to the same class. In parallel to this theory, we assume that our intuition model requires templates for the object classes, which can be utilized in the classification task of our model. In order to generate template images for each class, we use a popular image representation method based on the eigenvalue problem. In other words, for each object class, we estimate the eigen-images of the feature maps, obtained at the output of filters, in the convolution layer. We assume that these eigen-images represent our subconscious past experience.  
During the test step, we compare the filter outputs of a test image with the eigen-images of each class to estimate classes that resemble the test images the most. We use filter outputs of images from the training set belonging to the object classes to replace the filter outputs of the test images. This is how we attempt to complete the missing information of a test image.

For the generation  of the template images, the complete images  of the training set are fed as inputs to the  CNN model. A total of $M \times C$ feature maps are generated at the output of the  convolution filters,  for all classes $j=1,...,C$ and for all filters $k=1,...,M$.  We vectorize the feature maps of each image $\mathbf x_i$ by concatenating all the rows to generate  $1\times d$ dimensional $\mathbf{f_{ik}}$ feature vectors. Then,  we obtain a design matrix by concatenating  the feature vectors of all the images, for each class, $j$, and each filter, $k$, as follows:

\begin{equation}
P_{jk}^T = [\mathbf{f_{1k}}\ \mathbf{f_{2k}}\ ...\ \mathbf{f_{N_ck}}].
\end{equation}

Assuming that there are $N_c$ images in each class, the dimensions of our design matrice for a given class will be $N_c\times d$. We center the design matrices and form our $d\times d$ dimensional covariance matrix, $S_{jk}$, from the centered matrices as follows:

\begin{equation}
S_{jk} = P_{jk}^{T} P_{jk}.
\end{equation}

We obtain the eigenvalues and eigenvectors of our covariance matrix using the Eigenvalue Decomposition method as follows:
\begin{equation}
S_{jk} = \Psi_{jk}\lambda_{jk} \Psi^{T}_{jk}.
\end{equation}

In the equation above, $\Psi_{jk}$ represents a $d\times d$ matrix containing the eigenvectors  of $S_{jk}$, in its columns and $\lambda_{jk}$ represents a $d\times d$ matrix with eigenvalues corresponding to these eigenvectors on its diagonal. After empirically determining the highest $\delta\ll d$ eigenvalues, we define a $\delta\times d$ submatrix in the upper left corner of the $\Psi_{jk}$ matrix as a low-dimensional eigenvector matrix and describe it with the symbol $\psi_{jk}$. The columns of this submatrix,

\begin{equation}
    \psi_{jk} = [\mathbf{v}_1^{jk},..., \mathbf{v}_{\delta}^{jk}],
\end{equation}

\noindent correspond to the most important $\delta$ eigen-images of the feature maps, we estimated at the output of the convolution filters, during CNN training.

We assume that the  $1\times d$ dimensional eigen-images, $\mathbf{v}_q^{jk}$, for $q=1,...\delta$,  we estimated for each class, $j$  and for each filter, $k$, represent our subconscious memory, formed by our past experience. Thus, we create an additional layer to augment the CNN architecture, called  \textbf{memory layer}. 

The eigen-images, like the template images in the human mind, capture the differences among samples that belong to the same class. They also share the common properties which characterize the object class. With these characteristic properties, eigen-images represent the  template images in subconscious memory to formalize  the visual intuition to some extent.

\begin{figure}[t]
\centerline{\includegraphics[scale=0.1]{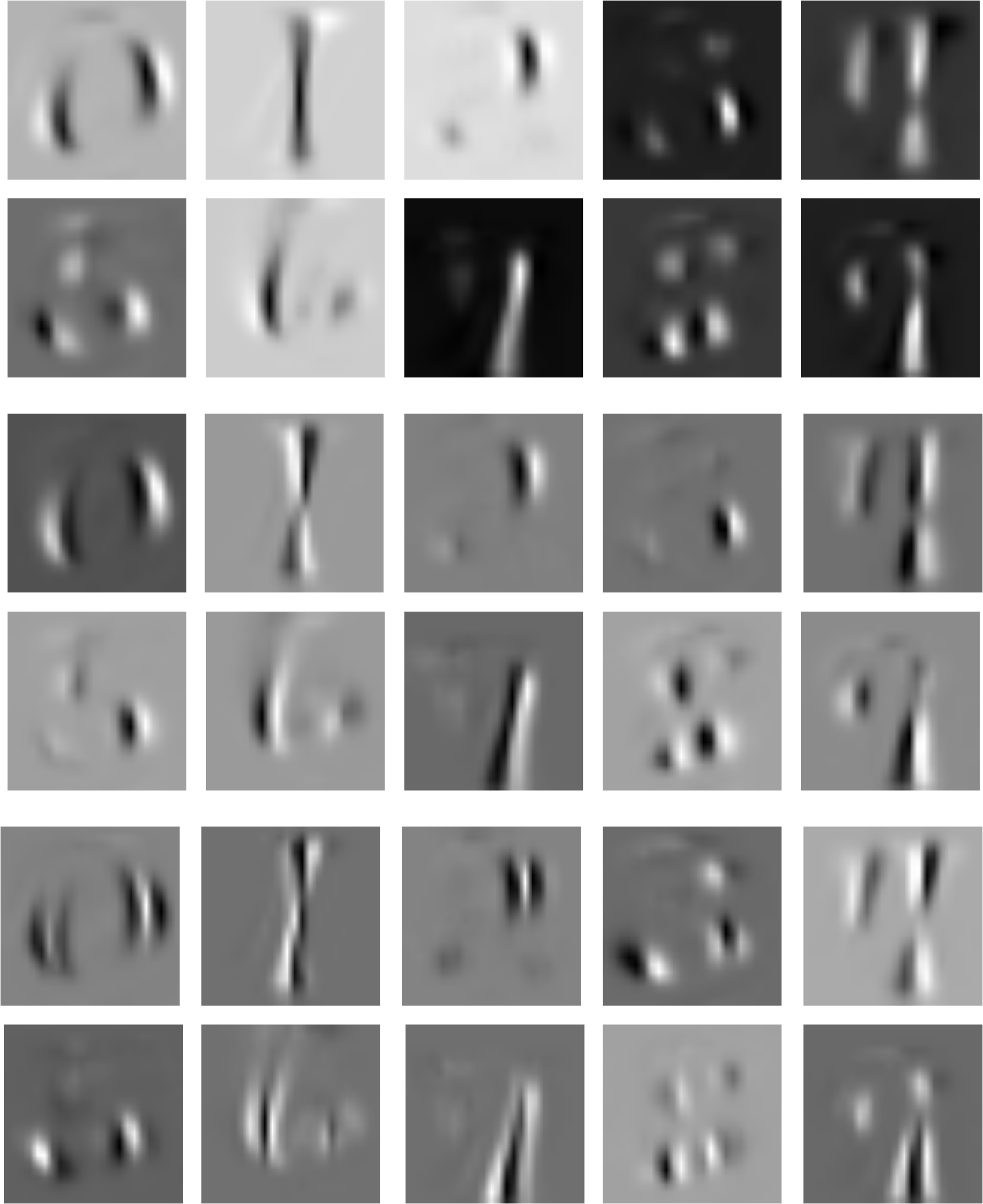}}
\caption{Eigen-images of number digits obtained at the output of  the first convolutional filter. The pairs of rows show the eigen images corresponding to the  first, second, and third largest  eigenvalues.  These eigen images are assumed to represent the object templates in subconscious memory.}
\label{fig_eigen}
\end{figure}

In addition to producing eigen-images, we store a set of feature maps in the memory layer. In this set, which we call \textbf{stock feature maps}, we record each feature map at the output of each filter, $k=1,...M$,  of a randomly selected image $\mathbf{x}_i$  from each $j$ class. Hence, the stock feature map dataset consists of  the feature maps of a single sample from each class with the corresponding label,

\begin{equation}
(\mathbf{f_{ik}},y_j) \in D^{stock},\quad \forall j.
\end{equation}

Our stock data set contains a total of $C\times M$ feature maps.

\subsection{Intuition layer: Completion of  Missing Information}

This layer, which can be considered the heart of our application, shows its effect by making changes on the feature maps estimated at the output of the convolution layer. Our model first attempts to recognize the test image fed into it with the intuition layer closed. If the accuracy of this image is below an empirically determined level, it restarts the recognition process by bringing the intuition layer into the open state. The intuition layer consists of two sub-layers, comparison of eigen-images and organizing feature maps, which works in parallel to the testing step of a CNN model.

\subsubsection{Finding the Dominant Class by Comparing  the Feature Maps of a Test image to the Eigen-Images}
In this subsection, we compare the feature maps of each test image to the eigen-images stored in the memory layer to estimate the class of the test image. Recall that the test image may consist of incomplete objects.

Formally speaking, for each test image, $\mathbf{x}_r\in D^{te}$, we obtain the feature map, $\mathbf {f_{rk}}$, for $k= 1,...M$, at the output of each filter of the trained CNN model. Thus, for each image, $\mathbf{x}_r$, we generate a total of $M$ feature maps. Then,  we compare each  feature map, $\mathbf{f_{rk}}$ with the eigen images obtained at the columns of the $\delta\times d$ dimensional eigen-image  matrix, $\psi_{jk}$, created on the memory layer, corresponding to each class, $j$.  The distance between the eigen-images and the feature maps of the test image is measured by  the Pearson Correlation defined below:

\begin{equation}
\rho(\mathbf{v}_{q}^{jk},\mathbf{f_{rk}}) = \frac{Cov(\mathbf{v}_{q}^{jk},\mathbf{f_{rk}})}{\sigma({\mathbf{v}_{q}^{jk}})\sigma({\mathbf{f_{rk}}})}.
\end{equation}

Finally, we calculate the highest Pearson value for each feature map, $\mathbf{f_{rk}}$, of the test image $\mathbf{x}_r$,
\begin{equation}
p_{k} = max_j(\rho(\mathbf{v}_{q}^{jk},\mathbf{f_{rk}})).
\end{equation}

The class label $j$, which gives us the maximum  Pearson value $p_k$ reveals that the feature map of filter $k$ is most similar to object class $j$. For a final decision about the  class label  of the test image, we use majority voting for all $p_{k}, k=1,...,M$, values. The class $j=J$ with the highest vote in all of the feature maps is called the \textbf{dominant class}.

\subsubsection{Replacing the Features Maps of the Incomplete Test Image by the Feature Maps of the Dominant Class}

Once we estimate the dominant class of our   test image, we replace the  feature maps of the incomplete test image  with the stock feature maps, which belong to the dominant class.

Formally speaking,  for each feature map, $\mathbf{f_{rk}}$, obtained from the image $\mathbf{x}_r \in D^{te}$, if $p_k \neq J$, then the feature map, $\mathbf{f_{rk}}$ is replaced with $\mathbf{f_{Jk}} \in D^{stok}$. With this process, we complete the missing information, based on the "past experience", mentioned in the Gestalt theory. 

\subsection{Improving the Testing  Accuracy With Subconscious Memory and Intuition layer}

The intuition layer, introduced in the previous section is kept off, while the prediction accuracy of the model is above an empirically determined threshold value. Once the model detects a low prediction accuracy, the intuition layer is activated and the model replaces the feature map of the test image to that of the corresponding stock image,  for the dominant class. This replacement increases the test accuracy, as we shall demonstrate in the next section.

Therefore, testing stage only differs from the testing stage of a model of a standard CNN architecture by the replacement of the feature map with the feature map of the stock image for the dominant class, when the intuition layer is on.  This replacement is expected to complete the missing information of the test image.

\section{Experimental Findings}

We employ the MNIST-digit set in our experiments.  Our model consists of  a simple CNN model with 2 convolution layers, each of which has a max pool layer. It is finalized by 3 fully connected layers. In the  convolution layers,  we estimate 6 filters of size 24x24. 

The augmented  CNN architecture includes \textbf{memory layer} and  \textbf{intuition layer}, which are activated when the model decides that there is no sufficient information in the test image to make a prediction.
The suggested intuition model operates on the outputs of the first convolution layer. 
\begin{figure}[t]
\centerline{\includegraphics[scale=0.7]{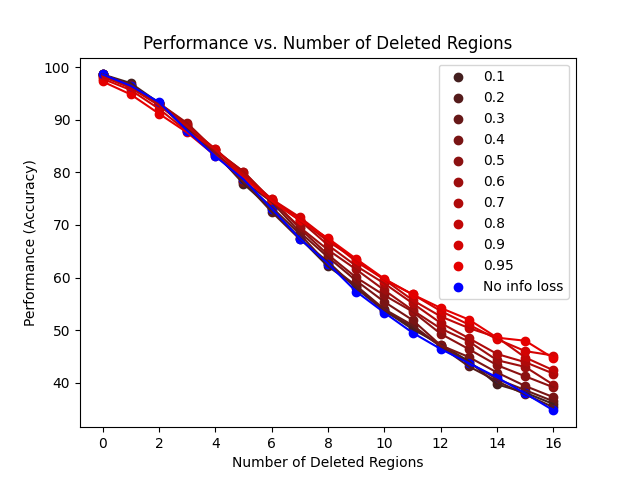}}
\caption{The blue  plot shows the performance of the standard CNN model.  Red  plots show the performance of the augmented CNN model for different threshold values. When the estimated class posterior is  less than a threshold $T$, the intuition layer is activated. As the shade of red gets lighter, the threshold $T$ for estimated class posterior  increases.}
\label{fig_comparison}
\end{figure}

\begin{table}[t]
    \renewcommand{\arraystretch}{1}
    \centering
    \caption{Performances of the augmented CNN model with threshold $T=0.9$. }
    \label{tab:results}
    \begin{tabular}{|c|c|c|}
    \hline
    Deleted Segments & No Intuition (\%) & Threshold: 0.9 (\%) \\
    \hline
    0 & 98.65 & 97.82 \\
    \hline
    1 & 96.48 & 95.68 \\
    \hline
    2 & 93.39 & 92.11 \\
    \hline
    3 & 87.97 & 87.82 \\
    \hline
    4 & 83.09 & 84.43 \\
    \hline
    5 & 78.69 & 79.13 \\
    \hline
    6 & 73.13 & 74.92 \\
    \hline
    7 & 67.40 & 71.48 \\
    \hline
    8 & 62.65 & 67.30 \\
    \hline
    9 & 57.34 & 63.03 \\
    \hline
    10 & 53.32 & 59.62 \\
    \hline
    11 & 49.50 & 56.81 \\
    \hline
    12 & 46.40 & 53.63 \\
    \hline
    13 & 43.71 & 50.03 \\
    \hline
    14 & 41.00 & 48.27 \\
    \hline
    15 & 37.94 & 46.00 \\
    \hline
    16 & 34.72 & 45.16 \\
    \hline
    \end{tabular}
\end{table}

\begin{figure}[t]
\centerline{\includegraphics[scale=0.25]{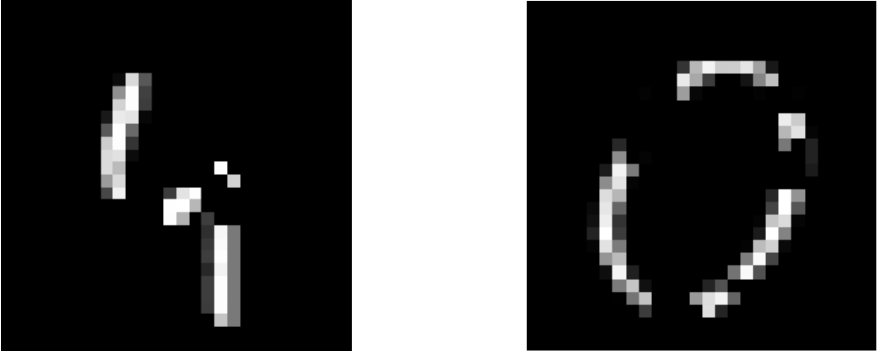}}
\caption{Incomplete digit images of 4 and 0, derived from the MNIST dataset with 8 segments removed.  The standard CNN model misclassifies digit 0 as 2 but correctly identifies digit 4. Conversely, the augmented CNN accurately labels 0 but classifies digit 4 as 9.}
\label{fig_int_lost_won}
\end{figure}

Before starting the experiments with the intuition layer, we generate a set of test datasets with incomplete images.  For this purpose, we segment the images in the test set. Then, we gradually remove the  $s=1,...,16$ segments, randomly for each test set. We test the performance of each incomplete test dataset. We  observe that as we delete more and more segments, the test performance decreases linearly (Figure \ref{fig_comparison}, blue plot). Finally, we feed the same test sets to our augmented CNN model with an activated intuition layer, using different threshold values ranging from $0.1$ to $0.95$. We illustrated the improved performance  on the same graph (Figure \ref{fig_comparison}, red plots). Notice that the suggested intuition model smoothly improves the performance of the test performance, compared to the classical CNN architecture, as we increase the number of deleted segments. The best performance of the augmented CNN is observed for the threshold  $T= 0.9$ . Hence, the augmented CNN activates the intuition module in the test step,  when the  posterior probability  of  the predicted class is less then $0.90\%$. The performance of different deleted number of segments $s$ are given in Table \ref{tab:results}. More than $10\%$ increase in accuracy can be achieved for higher information loss scenarios. 

In Figure \ref{fig_comparison} and Table I, it is observed that, when we delete segments with   $s<3$,  augmented CNN model slightly decreases the performance of the standard CNN. Hence, intuition layers should be activated, when the missing information is above a threshold.  On the other hand, we can observe that, when $s\ge 4$ the intuition layers  substantially  boost the performance as we increase $s$.  Figure \ref{fig_int_lost_won} shows two cases, where the intuition model corrects and  spoils the decision of the standard CNN model. The standard CNN model correctly classifies the  digit 4 of \ref{fig_int_lost_won} as 4. However, if we activate the intuition layers,  the model misclassifies it as 9. It is interesting to observe that humans may also perceive this incomplete image  as digit  9. On the other hand, without the intuition module, the model inaccurately identifies the digit  0 \ref{fig_int_lost_won} as 2. However, the augmented CNN model correctly labels it as digit  0. In both instances, the intuition module aligns more closely with our expectations based on Gestalt principles.

\section{Conclusion}

In this study, first, we show that CNN models are very error-prone in recognizing images with missing information. Based on this observation, we focused on overcoming the information deficit in classical CNNs. We made an effort to get over this with the inspiration from the human mind and visual intuition principles in Gestalt theory. In order to model visual intuition, we introduce two layers. In the first layer, we simulate the unconscious past experience of visual intuition by estimating a set of eigen-images of the learned representation space, obtained at the output of the convolution layer of a CNN model. In the second layer, we define a set of rules to replace the incomplete learned representation of a test image with a complete representation of an image in a stock image dataset.

Our experimental results show that augmenting a CNN architecture with the suggested visual intuition layers improves the robustness to missing information. We hope that our study can be a starting point for triggering future ideas on modeling intuition in machine learning literature.

\vfill\pagebreak
\bibliographystyle{IEEEbib}
\bibliography{refs}

\end{document}